\documentclass[journal, 12pt, onecolumn, draftclsnofoot]{IEEEtran}

\ifCLASSINFOpdf
\else
\fi
\usepackage{rotating}
\usepackage{pdflscape}
\usepackage{graphicx}

\begin{document}
%

\title{Which Facial Expressions Can Reveal Your Gender? A Study With 3D Faces}

\author{Baiqiang~XIA
\thanks{Baiqiang XIA is with the School of Computing and Communications, Lancaster University, UK, e-mail: b.xia@lancaster.ac.uk. Manuscript created May 1, 2018.}}
 
\markboth{In Submission to IEEE Transactions on Information Forensics and Security}{}

\maketitle

\begin{abstract}

Human exhibit rich gender cues in both appearance and behavior. In computer vision domain, gender recognition from facial appearance have been extensively studied, while facial behavior based gender recognition studies remain rare. In this work, we first demonstrate that facial expressions influence the gender patterns presented in 3D face, and gender recognition performance increases when training and testing within the same expression. In further, we design experiments which directly extract the morphological changes resulted from facial expressions as features, for expression-based gender recognition. Experimental results demonstrate that gender can be recognized with considerable accuracy in Happy and Disgust expressions, while Surprise and Sad expressions do not convey much gender related information. This is the first work in the literature which investigates expression-based gender classification with 3D faces, and reveals the strength of gender patterns incorporated in different types of expressions, namely the Happy, the Disgust, the Surprise and the Sad expressions.

\end{abstract}


\IEEEpeerreviewmaketitle

\section{Introduction}

\subsection{Gender recognition from facial appearance}

Gender recognition plays an important role in human interactions. We human beings can recognize gender naturally and effectively \cite{han2015,gilani2014}. In recent years, automated gender classification has attracted many attentions in computer vision domain, facilitating applications in video surveillance, human computer-interaction, anonymous customized advertisement and image retrieval \cite{dantcheva2011, han2015, sun2018}. The first automated gender classification approach was proposed in the early 1990's \cite{golomb1990}. Following that, 2D faces has been extensively researched for gender recognition, as surveyed in \cite{sun2018}. With the development of 3D sensing technology, the recent decade has viewed a rapid growth of 3D based gender recognition approaches. Compared to the 2D counterpart, 3D face models can capture the real facial shape without 2D projection, and are invariant to face/camera pose and illumination changes. With 3D facial models, comparable or even higher gender recognition performances have been reported \cite{xia2013,huang2014}. 

The introduction of the FRGCv2 dataset \cite{phillips2005} has greatly boosted the 3D face related researches, e.g 3D face recognition, facial expression recognition and gender/age/ethnicity recognition. For 3D face based gender recognition, many approaches have been reported with the FRGCv2 dataset, which can be summarized into two categories according to their feature engineering methodologies: \textit{(i)} the works that extract image-based features like the wavelets \cite{Toderici2010}, the Local Binary Patterns (LBP) \cite{wang2013,xia2013} or LBP variant \cite{huang2014}, and the Shape Index \cite{wang2013}; and \textit{(ii)} the works which extract features from 3D facial meshes, such as the needle maps \cite{Wu2010}, the selected facial curves \cite{ballihi2012}, the facial landmark distances \cite{gilani2013}, the Dense Scalar Field (DSF) features \cite{xia2015}, and the Correspondence Vectors (CVs) \cite{tokola2015}. These works have demonstrated consistently that gender can be effectiveness recognized using 3D face models. Whereas, as the 3D mesh based approaches preserve the true facial geometry, more fine-grained studies can be further addressed, such as revealing the exact locations of salient features on 3D facial surfaces in the context of gender classification \cite{gilani2014,xia2015,xia2017}.



\subsection{Gender recognition from body and face dynamics}

Apart from the static appearances, human also present gender in their body and facial dynamics. The study of gender recognition from motion originated from the work of \cite{kozlowski1977}, who found that the gender of walkers could be identified from point-light displays (lights attached to the major body joints). Hip sway often indicates female motion and shoulder movement indicates male motion \cite{johnson2005}. Females present excessive nodding, blinking and overall amount of movement, which helps their gender recognition \cite{morrison2007}. Recent studies further demonstrated that the motion based gender perception is not affected by the gender of the motion presenter \cite{Zibrek2015}, and motion related cues were found to be separable from morphological cues for gender perception \cite{zibrek2013,Zibrek2015}. 

Besides the body movements, facial movements contribute even more effectively to the perceived femininity and masculinity. In comparison with head motion, facial movements incorporate more gender related cues \cite{hill2001}. Raising the eyebrows makes the face more feminine, while lowering the eyebrows increases the facial masculinity \cite{campbell1999}, which consists with the fact that females tend to have a larger brow-lid distance than males. \cite{hill2001, hill2003, morrison2007} found that gender can be effectively recognized from facial motions by human observers. Reports have also shown consistently that emotional bias on gender perception exists world-widely crossing countries and cultures, where the angry motions are perceived as more masculine and sad motions are perceived as more feminine \cite{fischer2004,johnson2005,Zibrek2015}, regardless of the actual gender of the motions performer \cite{johnson2005, Zibrek2015}. 
It's reported that Anger makes harder the gender recognition from female motions, Neutral and Sad enhance the gender recognition from the female motions, and Happy and Fear increase the accuracy of gender recognition from both male and female motions. 

\begin{figure*}
\centering
\includegraphics[width = .8\linewidth]{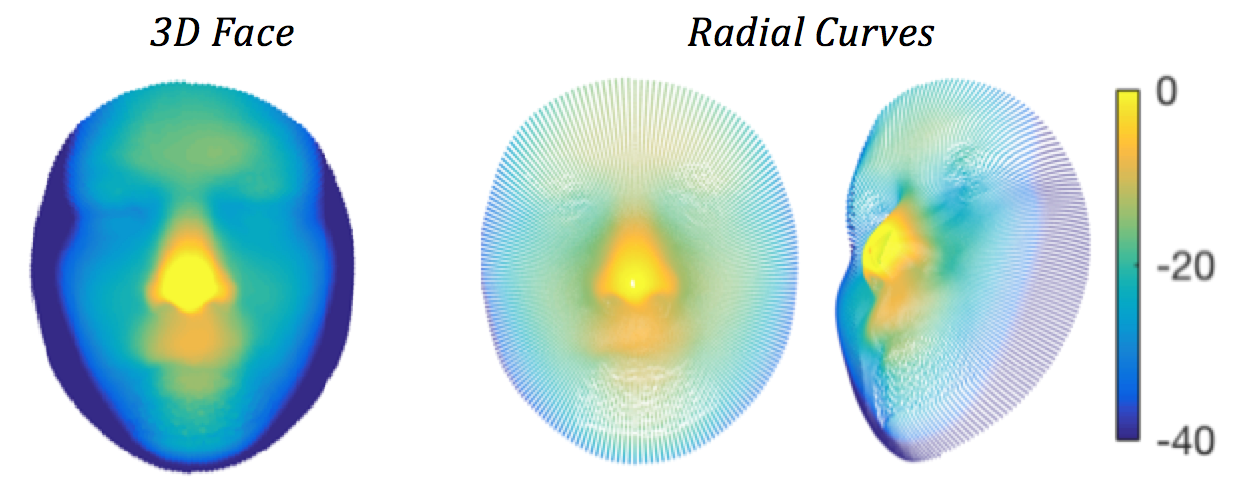}
\caption{Preprocessed 3D face represented by a set of radial curves emanating from nosetip. This representation enables shape comparison between any two 3D faces, on each point of the indexed radial curves. The colormap is with the unit of millimeter (mm).}
\label{fig:radial_curves}
\end{figure*}

\subsection{Motivation to this study}

The related works present clearly the capability and advantage of the 3D face modality in gender recognition, the strong interaction between facial movements and the perceived gender, and the feasibility of gender recognition from facial motions (as revealed by the performances of human observers). However, in computer vision research, however, gender recognition from facial expressions has received very limited attention. Existing studies \cite{bilinski2016, dantcheva2017} explored only the facial smile-dynamics captured in 2D videos, where \cite{bilinski2016} used fisher vector embedding techniques with five different image-based descriptors, and \cite{dantcheva2017} extracted 2D facial landmarks from faces, with underlying assumption that faces are frontal. Arguably, these 2D imagery based works have incorporated inaccuracies resulted from the camera/head pose changes and illumination changes, and the facial geometry has been distorted due to 2D projection of the original 3D face. Also, many other facial expressions, such as the Disgust, the Surprise, and the Sad expressions, were not explored in their studies. Whereas, it has been revealed with human observers that different expressions have different effectiveness in revealing gender \cite{fischer2004,johnson2005,Zibrek2015, zibrek2013,Zibrek2015}. Considering these aspects, we are motivated to establish the first study with 3D face models for expression based gender recognition, where we extract directly the facial morphological changes that happen naturally in 3D, and examine the effectiveness of expression-based gender recognition in four typical expressions, namely the Happy, the Disgust, the Surprise and the Sad expressions. 

The remaining contents of this work are organized as follows: we present the methodology and summarize the main contributions in section 2; Section 3 reports the expression-aware gender recognition results. Expression-based gender recognition performances are detailed in Section 4; Section 5 concludes the work and discusses the limitations and future research perspectives.

\section{Methodology}

In our work, following a typical preprocessing pipeline, the raw 3D faces are firstly processed with hole filling, central facial part cropping, mesh smoothing, and ICP-based face frontalization. Then the preprocessed 3D faces are approximated by a set of indexed radial curves emanating from the nosetip, as shown in Figure \ref{fig:radial_curves}. The radial curves representation has been widely adopted in 3D face studies \cite{ballihi2012, xia2015}. In effect, it sets a framework for shape comparison between two 3D faces, which can be achieved by analyzing pair-wisely the indexed curves at each indexed point. In the next, the depth values at each indexed point of the radial curves are taken as features in our gender recognition studies. In this work, two types of studies are established, namely: \textit{(i)} the expression-aware gender recognition study, which aims at revealing whether gender recognition performance will increase when training and testing on instances from a specific expression, than training generally on all available instances; \textit{(ii)} the expression-based gender recognition study,  which employs the geometric facial changes during expression for gender recognition, to reveal the strength of gender patterns in expressions. The authors note that the purpose of this work is not to improve the 3D face based gender recognition performance in the state of the art, but to reveal the impact of considering facial expressions in 3D gender recognition, and the effectiveness of using 3D facial expression features alone for gender recognition. The main contributions of this work are summarized as follows:

\begin{itemize}

\item This is the first work investigating the impact of facial expressions for gender recognition with 3D faces. To this end, we demonstrate that facial expressions add discriminative power to 3D gender recognition, through the comparison between expression-general and expression-specific gender recognition experiments.

\item This is the first work exploring gender recognition from facial expression with 3D face models, which covers four typical facial expressions, namely Happy, Disgust, Surprise, and Sad. 

\item Experimental results demonstrate that Happy and Disgust rank the first two most discriminative expressions for gender recognition. The Surprise and Disgust expressions contribute little to gender recognition.

\end{itemize}

In the next, we present the experimental evaluations concerning the above studies.

\section{Expression-Aware Gender Recognition}

\subsection{3D Dataset Explained}

\begin{figure}
\centering
\includegraphics[width = \linewidth]{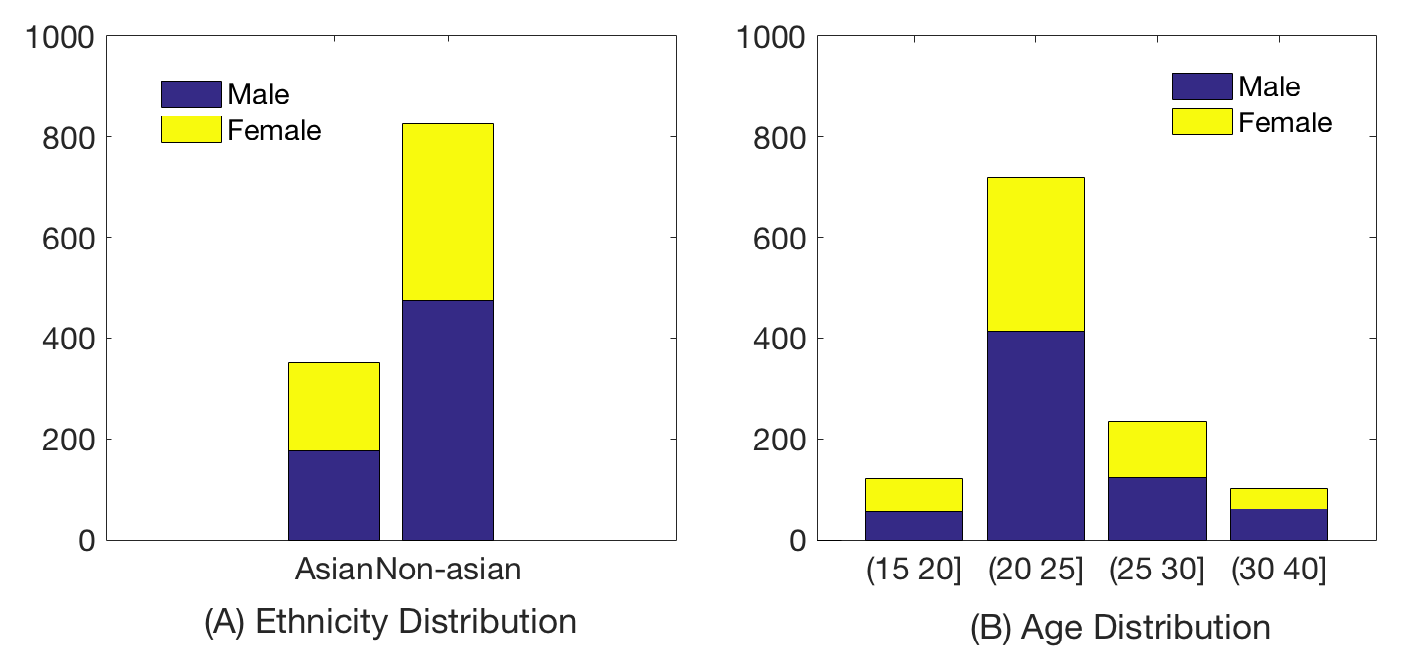}
\caption{Ethnicity and Age distributions of the experimented data subset.}
\label{fig:demographics}
\end{figure}

The FRGCv2 database was collected by researchers from the University of Notre Dame \cite{phillips2005} and contains 4007 3D face scans of 466 subjects with differences in gender, ethnicity, age and facial expressions. About 60\% of the faces have a neutral expression, and the remaining scans show expressions of happiness, sadness, surprise and disgust. In our experiments, we remain only the subjects which have at least one neutral 3D face scan and an expressive 3D face scan, and with age no more than 40. In further, we take only the first scan of each subject concerning each expression, which results into 1176 scans from 324 persons. For gender, 142 are female and 182 are male.  The Male and Female subjects have insignificant difference in terms of Ethnicity (t-test between Asian and Non-Asian: $p = 0.45 >> 0.05$ ) and Age (t-test between the males age and females age: $p = 0.52 >> 0.05$). The distributions of Ethnicity and Age concerning each gender are presented in Figure \ref{fig:demographics}. Male and female scans are balanced in each Ethnicity and Age groups. In this case, we are able to study the effect of facial expressions on gender recognition alone, without being confounded by other co-existing demographic traits, such as the Ethnicity and the Age of subjects. In our experiments, we extract densely 100 indexed radial curves per face, and each facial radial curve is described with 40 indexed points. With the depth information extracted at each indexed point, a 4000 dimensional feature vector is formed to represent each 3D face. For gender classification, the Linear-kernel Support Vector Machine (SVM) \cite{suykens1999} and the Random Forest \cite{breiman2001} are employed as classifiers.

\subsection{Expression-general Gender Recognition}

With the data subset, we first perform subject independent gender classification with a Leave-One-Out (LOO) cross-validation setting. In each round, we holdout one subject for testing, and all the other subjects for training the classifier. We perform these experiments to demonstrate the effectiveness of the employed gender classification methodology, which would form the basis for further analysis. If the gender classification approach doesn't work effectively (e.g with accuracy close to random guess), no further analysis can be addressed to reveal the potentially influential factors, such as facial expressions. 

Shown in Table \ref{tab:gender_general}, the LOO cross validation experiments yield a gender classification rate of 90.99\% with linear-kernel SVM classifier, and 89.03\% with the 100-trees random forest classifier. These results demonstrate the effectiveness of using the depth information from 3D scans for gender recognition. Recall that the SVM and Random forest classifiers have very different learning strategies to make classification. The SVM projects features into higher dimensional hyperspace and then takes the hyperplane which optimally separates the training instances as reference to make predictions on new instances, while the random Forest classifier creates an ensemble of random decision trees and summarizes their outputs to make a new decision. With the effective gender recognition performance with both classifiers, it demonstrates that that the gender classification effectiveness is not due to the choice of a particular learning strategy, but is more resulted from the informative features we have employed.

The above results demonstrate the effectiveness of gender recognition using depth information, in terms of the \textit{\textbf{average classification rate}} which underlies that gender perception is a binary classification problem. This underlying ideology probably comes from the biological notion of gender, which is strictly determined into binary classes by the chromosomes and stays intact throughout individual's life. In construction of computer vision datasets, the gender ground truth also usually refers to the self-reported biological gender. However, the differences between \textit{\textbf{biological gender}} and \textit{\textbf{apparent gender}} in vision should be better discussed, although they usually comply with each other. Compared with the biological gender, the apparent gender is presented in less determinative and even highly changeable external cues, such as the appearances, the morphology, the behavior, and the clothing and accessories. In extreme cases, such as gender disguise, they can be totally the opposite. When recognizing gender from apparent cues, the observer is actually rating the instance, to be more masculine or more feminine. The apparent gender should be considered more as a magnitude of masculinity or femininity, than merely two discrete states. In studies with human observers, some researches have pioneered to address apparent gender recognition as a multi-level rating \cite{zibrek2013,Zibrek2015} problem or a scoring problem \cite{gilani2013}. In computer vision domain, 
\cite{gilani2013} also found that the scores from LDA classifier trained with a set of facial landmarked based features correlate strongly with the subjective gender scores from human observers. \cite{xia2014} took the voting ratio of the trained decision trees as magnitude of sexual dimorphism and achieved higher gender classification accuracy. These studies suggest that, in the experimental evaluations, we should go deeper to examine
the classifiers' output than merely reporting the classification rate.

\begin{table}[]
\centering
\caption{Expression-general gender recognition. SVM: Linear-kernel Support Vector Machine, RF: 100-trees Random Forest.}
\vspace{0.5cm}
\label{tab:gender_all}
\begin{tabular}{lllll}
\hline
\multicolumn{1}{|l|}{} & \multicolumn{1}{l|}{Female} & \multicolumn{1}{l|}{Male} & \multicolumn{1}{l|}{ALL}   \\ 
\hline
\multicolumn{1}{|l|}{SVM} & \multicolumn{1}{l|}{90.66\%} & \multicolumn{1}{l|}{91.24\%} & \multicolumn{1}{l|}{90.99\%}   \\ 
\hline
\multicolumn{1}{|l|}{RF} & \multicolumn{1}{l|}{88.17\%} & \multicolumn{1}{l|}{89.72\%} & \multicolumn{1}{l|}{89.03\%}   \\ 
\hline
\multicolumn{1}{|l|}{$\sharp$ Scans} & \multicolumn{1}{l|}{524} & \multicolumn{1}{l|}{652} & \multicolumn{1}{l|}{1176}   \\
\hline

\end{tabular}
\label{tab:gender_general}
\end{table}

\begin{figure*}
\centering
\includegraphics[width = \linewidth]{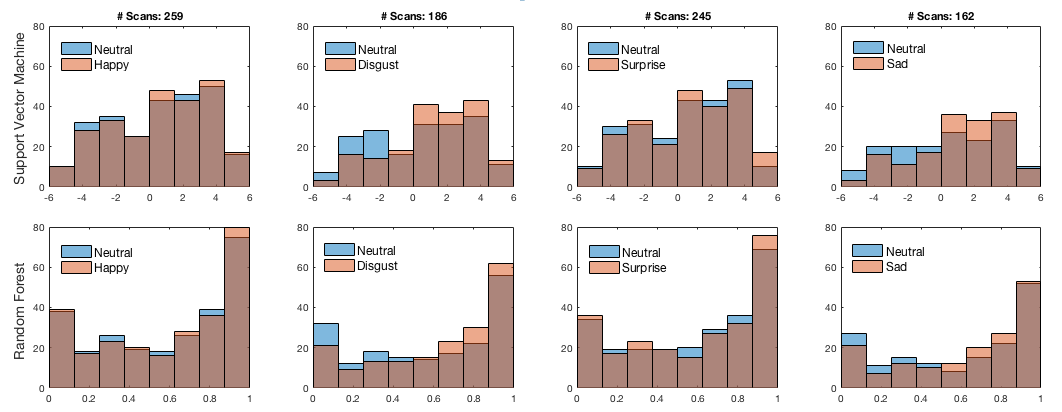}
\caption{Histograms of the critical values leading to classification decisions in SVM and Random Forest, for neutral and expressive faces of same subject. For SVM, when the instance's distance to the learned hyperplane is negative, the gender is judged as male, otherwise female. In Random Forest, when the voting ratio (for female) from the trees is below 0.5, the gender is taken as male, other wise female. These plots demonstrate that facial expressions influence the gender perception in machine vision. Especially with the Disgust and Sad expressions, the faces are perceived as less masculine by both classifiers.}
\label{fig:decision_expression}
\end{figure*}

The above analysis suggests that there are more detailed gender knowledge in the classifiers' output beyond the classification rate. Considering this, we are motivated to make use of the critical values resulting into the classifications in each classifier. Recall that in SVM, the classification decision is made according to the \textit{instance's distance} to the learned hyperplane. In Random forest, the decision is achieved by the \textit{voting ratio} of the random trees. To reveal the influence of expression in gender recognition, we look into these critical values group-wisely for each facial expression, and compare the corresponding critical values on the neutral faces from same subjects. Figure \ref{fig:decision_expression} plots the critical values for SVM and Random Forest classifiers in the LOO cross-validation experiments, between neutral faces and expressive faces. It shows that the apparent gender perception of the classifiers is influenced by facial expressions. Especially with the Disgust and Sad expressions, the faces are perceived as less masculine than in their neutral scans. 

\subsection{Expression-specific Gender Recognition}

The fact that the perceived gender receives influences from facial expressions suggests that expression variation should be addressed properly in gender recognition studies. As faces of different expressions exhibit relatively different gender patterns, gender recognition models trained with faces in a specific facial expression may not generalize well on faces in another expression. In this case, we are motivated to perform a set of expression-specific experiments, which train on faces with a specific expression, and test with the same or another expression. Intuitively, when training and testing within the same expression, a higher recognition accuracy is expected. While when training and testing cross expressions, the related accuracy should decrease. 

\begin{figure}
\centering
\includegraphics[width = 1\linewidth]{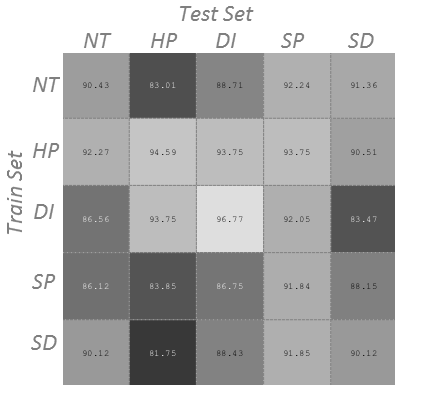}
\caption{Expression-specific gender classification results. For the cells, the brighter the color, the higher the gender recognition accuracy. NT: Neutral, HP: Happy, DI: Disgust, SP: Surprise, SD: Sad }
\label{fig:expression_aware_gender_classificaiton}
\end{figure}

In Figure \ref{fig:expression_aware_gender_classificaiton}, we presents the expression-specific gender recognition results. In general, the gender recognition models generalize well even testing faces with another expression, which means that the 3D faces can preserve most of the gender related patterns even with facial expression changes. While, when training and testing within Happy and Disgust, Surprise scans, the recognition accuracies are considerably higher than training with other expressions. This means the trained models have been tuned to adapt to the particular expression. When summarizing the results training and testing with the same expression (results in the diagonal cells), a gender recognition accuracy of 92.60\% is achieved, which is 1.60\% higher than in the expression-general setting. Also, here the amount of training data is just about 1/5 of the size than training on the whole data subset in the expression-general setting, which could have potentially limited the performance of expression-specific gender recognition. In summary, these results demonstrate that gender pattern exists in an expression-specific way, and recognizing gender within each expression can effectively increase the gender recognition performance. 

In addition, we find the model trained on Happy subset has considerably good generalization property, with 93.00\% mean gender classification rate (by summarizing the 2nd row in Figure \ref{fig:expression_aware_gender_classificaiton}). It demonstrates that classification models trained with happy faces are of better generalization ability for gender recognition. This finding consists with \cite{zibrek2013}, which found that with the Happy expression human observers performed better gender recognition on both males and females.

\subsection{Discussions}

Facial expressions change the textural and morphological appearance of a face. With human observers, it has been reported that gender perception and expression perception from 3D face are correlated with each other \cite{zibrek2013}. With the results presented above, we demonstrate that facial expressions have also a considerable effect on the perceived gender, within the machine vision context. Facial expressions can provide additional information beyond facial morphology for gender prediction. With this, we are motivated to carry on studies which reveal the strength of facial expressions patterns (apart from the face morphology) in gender recognition, in the next section.

\section{Gender recognition from facial expressions}

\subsection{3D Facial Expression-based Gender Recognition}

\begin{figure*}
\centering
\includegraphics[width =  \linewidth]{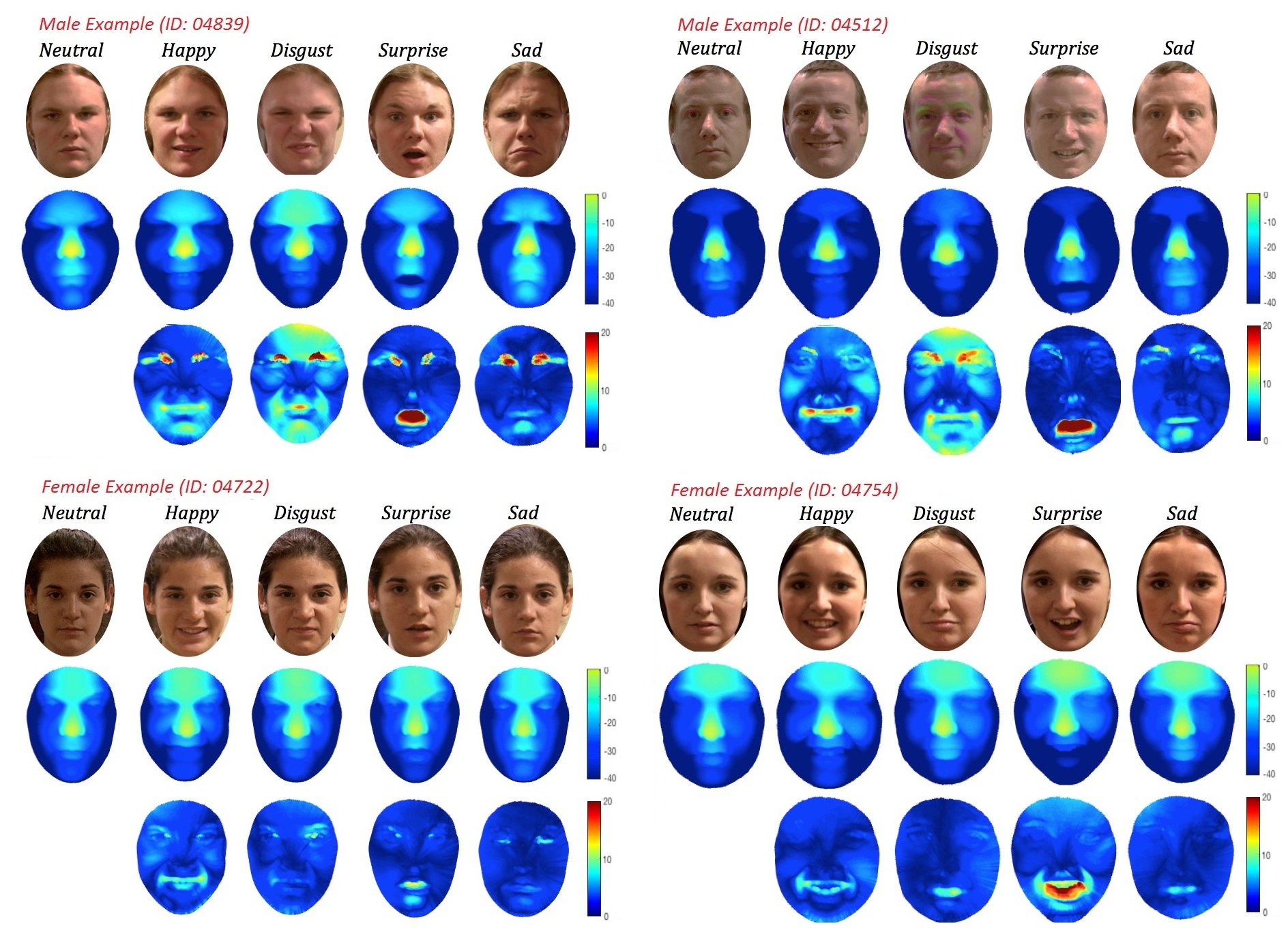}
\caption{Male and Female Faces Examples and Extracted Features. For each participant, three rows of plots are presented, respectively: (i) the upper row shows the 2D facial scans with different facial expressions; (ii) the middle row visualizes the corresponding 3D faces; and (iii) the last row presents the absolute values of extracted features (point-wise difference between neutral 3D face and expressive face from the same person) as colormap on faces. Visually, male faces present more facial geometric changes than female faces during facial expressions, especially in the Happy and Disgust expressions.}
\label{fig:feature_illustration}
\end{figure*}

An expressive face can be considered as the composite of the neutral face which capture the default facial morphology, and the morphological changes resulted from facial expressions. To reveal the strength of expression-based gender recognition, we should first remove the static morphological cues, and remain only the features resulted from facial dynamics. Thanks to the advantage of 3D face models in preserving the true 3D facial morphology, in this study, we can obtain the dynamic features by simply subtracting each expressive face with the corresponding neutral scan from the same subject. This captures the morphological changes between expressive faces and their neutral shapes. Figure \ref{fig:feature_illustration} depicts the resulted facial dynamic features in each facial expression as colormap on the faces, where both male and female examples are provided. In general, the male examples exhibit much larger facial deformations with facial expressions, especially with the Happy and Disgust expressions. In happy expression, the deformations around the mouth, the cheeks and the eyes in male faces are much larger than in the female faces. In Disgust expression, the male faces exhibit large deformations all over the faces, while the female faces present relative subtle changes around the mouth and eyes regions. This observation echoes \cite{fischer2004} where males were found to show more powerfulness in emotions than females, and also the finding in \cite{deng2016} where males demonstrated more intense emotional experiences than females. In contrast, facial deformations are much lower in magnitude in the Surprise and Sad expressions. As expected, the Surprise expression mainly exhibits within the mouth region (mouth opened) and eyebrow regions (eyebrows lifted), and the Sad expression mainly shows signs in the mouth and eye regions. The differences between males and females are not as evident as in the Happy and Disgust expressions.

With the extracted facial expression features, a set of Leave-One-Person (LOO) out experiments are performed using the linear-kernel SVM classifier. The results are presented in Table \ref{tab:exp-based-gender-recognition}. With the facial motion features from the Happy, Disgust, Surprise and Sad expressions, we achieve gender classification rate of 79.15\%, 72.04\%, 65.13\% and 66.05\% with the corresponding data subset. These results demonstrate that facial expressions, especially the Happy and Disgust expressions, contain rich cues for gender perception. These results echo the Figure \ref{fig:expression_aware_gender_classificaiton}, where the gender classification rates increased most significantly when training and testing within the Happy and Disgust expressions. In contrast, the gender recognition performances with Surprise and Sad expressions are much lower. Interestingly, research from another aspect has shown very similar observations \cite{deng2016}. When watching videos that induce anger, amusement, and pleasure, men often have more intense emotional experiences characterized by larger decreases in Heart Rate, and when watching videos that induced horror and disgust, women reported lower valence, higher arousal, and stronger avoidance motivation than did men. Also, no gender difference was observed in induced sadness or surprise with the Heart Rate and Overt Report measurements. In summary, it means Anger, Happy and Disgust have very meaningful gender related differences, while Surprise and Sad incorporate little gender information.

\begin{table}[]
\centering
\caption{Gender recognition from 3D facial expressions}
\label{tab:exp-based-gender-recognition}
\vspace{0.5cm}

\begin{tabular}{lllll}
\hline
\multicolumn{1}{|l|}{Expression} & \multicolumn{1}{l|}{Female} & \multicolumn{1}{l|}{Male} & \multicolumn{1}{l|}{All} & \multicolumn{1}{l|}{$\sharp$ Scans} \\ \hline
\multicolumn{1}{|l|}{Happy} & \multicolumn{1}{l|}{74.44\%} & \multicolumn{1}{l|}{82.01\%} & \multicolumn{1}{l|}{79.15\%} & \multicolumn{1}{l|}{259} \\ \hline
\multicolumn{1}{|l|}{Disgust} & \multicolumn{1}{l|}{67.50\%} & \multicolumn{1}{l|}{75.47\%} & \multicolumn{1}{l|}{72.04\%} & \multicolumn{1}{l|}{186} \\ \hline
\multicolumn{1}{|l|}{Surprise} & \multicolumn{1}{l|}{63.48\%} & \multicolumn{1}{l|}{66.92\%} & \multicolumn{1}{l|}{65.31\% } & \multicolumn{1}{l|}{245} \\ \hline
\multicolumn{1}{|l|}{Sad} & \multicolumn{1}{l|}{65.16\%} & \multicolumn{1}{l|}{67.12\%} & \multicolumn{1}{l|}{66.05\%} & \multicolumn{1}{l|}{162} \\ \hline
\end{tabular}
\end{table}

To better understand the gender related facial expression differences, we investigate the features in two ways to provide insights of the experimental results, namely: \textit{(i)} a statistical t-test based analysis on the features, to reveal accurately the locations of the informative features in 3D face for distinguishing gender, and \textit{(ii)} a Principle Component Analysis (PCA) on the features, concerning each gender and each expression, to show the feature difference between males and females in each expression in a summarized way.

\begin{figure}
\centering
\includegraphics[width = \linewidth]{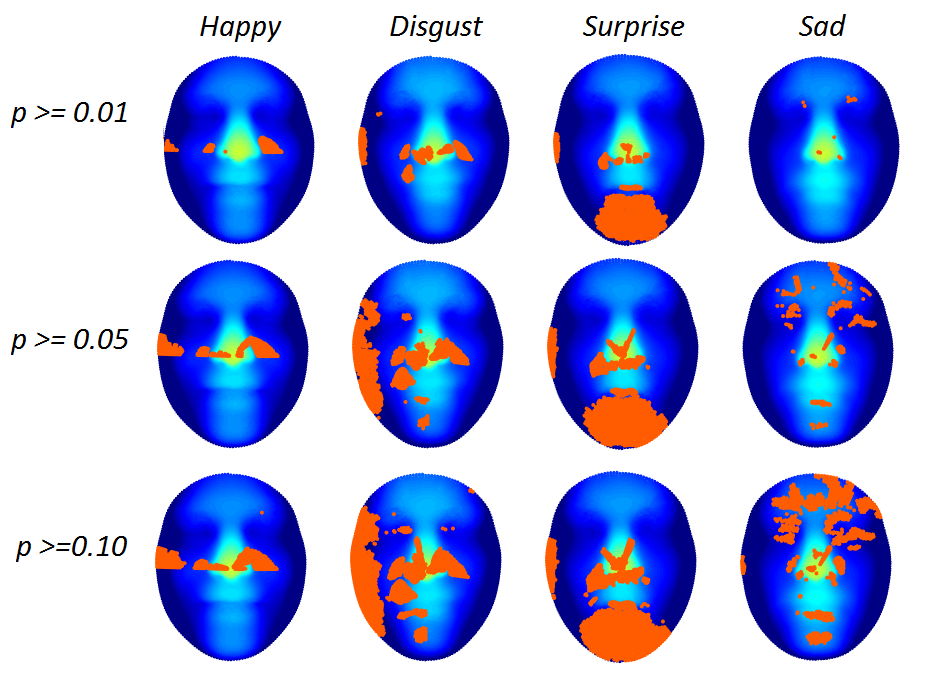}
\caption{Location of salient features revealed by t-test where males and females have different feature means. When the difference is observed at a certain significant level (p = 0.01, 0.05 or 0.10), the corresponding point is marked in red. Clearly, in different facial expression, the salient regions for inferring gender are different.}
\label{fig: t_test}
\end{figure}

With the statistical t-test analysis, we aims at revealing the distributional differences between males and females with each feature. Figure \ref{fig: t_test} presents the t-test results of features as colormap on each facial point. Given a significance level $p$, when significant difference exists on the feature's mean between males and females, the corresponding facial point is marked in red. Clearly, informative features locate at different facial regions in different facial expressions. In Happy expression, the most salient features locates in the nose and cheek region. In Disgust expression, salient features locates around the nose, and the right cheek. The mouth and nose regions are highlighted in Surprise expression, and the Sad expression exhibits very few high-saliency features ($p > 0.01$) for gender recognition.

\begin{figure*}
\centering
\includegraphics[width = \linewidth]{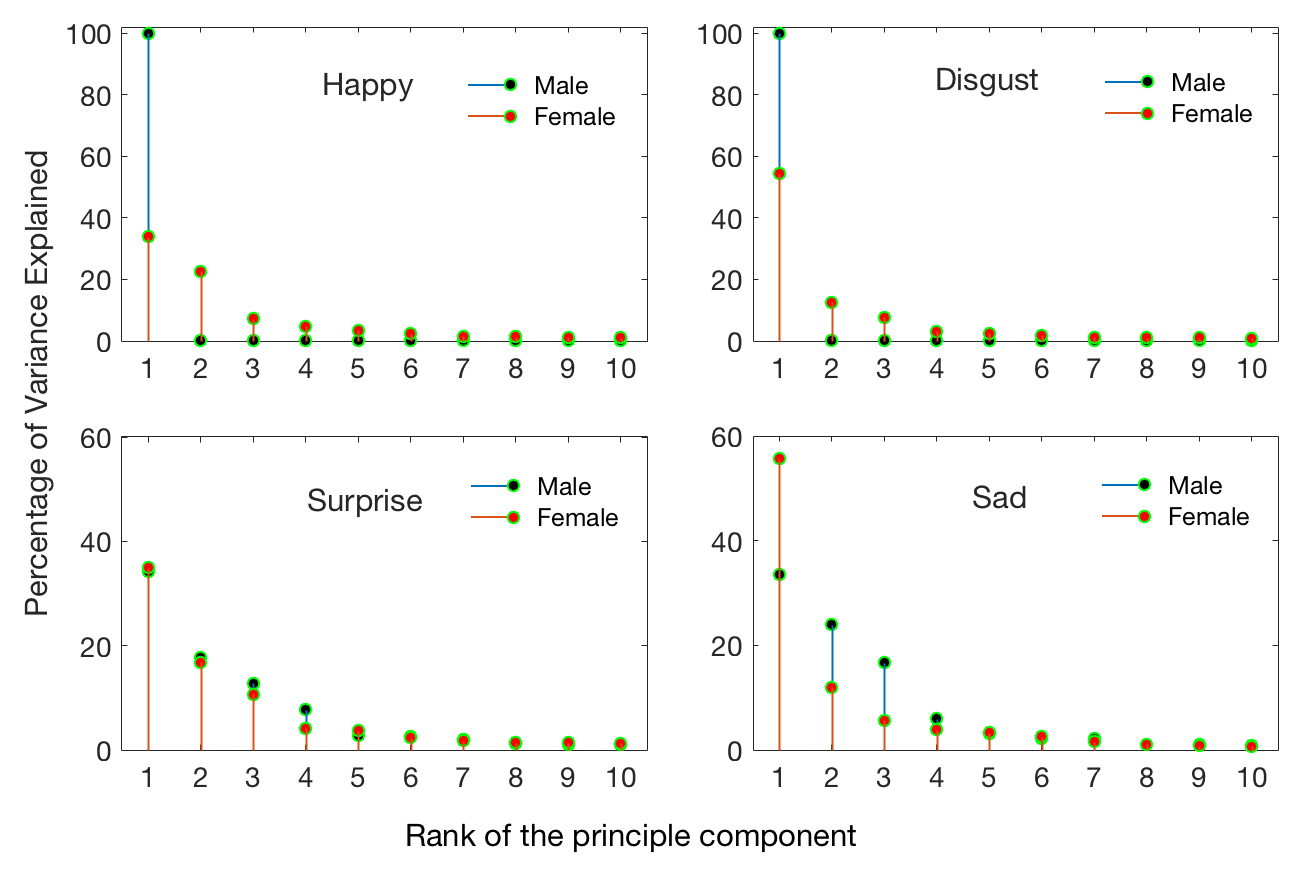}
\caption{Per expression and gender PCA analysis on the extracted features. In each plot, the x-axis signifies the rank of the principle component, and the y-axis presents the percentage of original feature variance explained by the principle component. Happy and Disgust expressions in males are explained dominantly by the first principle component ($>99\%$), while the variance in female features need several principle components to explain. The Surprise and Sad expressions are characterized by several principle components, in both gender.}
\label{fig:pca_per_gender_exp}
\end{figure*}

The idea behind PCA analysis is to capture effectively the variances presented in high dimensional features with a lower dimensional subspace composed of principle components. The PCA analysis results are presented in Figure \ref{fig:pca_per_gender_exp}, concerning each principle component and its power in explaining the variance presented in the original features. The plots show vividly that the feature variance in Happy and Disgust expressions are dominantly explained by its first principle component in the males ($>99\%$), while for females more principle components are needed to explain the feature variances. For the Surprise and Sad expressions, the feature variances are explained by a combination of several principle components, in both males and females. The gender related differences are not as obvious as in the cases of Happy and Disgust expressions. These results offer a perspective to explain the relatively higher gender recognition performances achieved in Happy and Disgust expressions, and the relatively lower performances with the Surprise and Sad expressions. 

\subsection{Comparative study with 2D landmarks}

In a similar way, we explore performance of gender recognition with the corresponding 2D RGB facial images provided in the same dataset. These color images are strictly in pair with the 3D face scans, as they were captured together and simultaneously with the same camera device. The original resolution of the images is $640 * 480$, capturing both the background and the face. With this 2D dataset, we first detect automatically the facial landmarks using the method proposed in \cite{kazemi2014}. The resulted 68 landmark positions represent the face in a sparse way, covering coarsely the regions of eyebrows, eyes, nose, mouth and lower face contour. Then we examine the 2D landmarks in two ways for gender recognition: \textit{(i)} we align all the facial landmarks with the nose-tip landmark, and vectorize all the coordinate values to form a 136 dimensional feature vector for each face; \textit{(ii)} we calculate the distance between each pair of landmarks and result in a 2278 dimensional feature vector for each face. With the same Leave-One-Out cross-validation experimental setting, both type of features are revealed to be effective in gender recognition, as shown in Table \ref{tab:gender_all_2D}, although the performances are comparatively lower than using the 3D face ( shown in Table \ref{tab:gender_all}: 99.09\% ).

In further, we perform expression based gender recognition using the feature differences between expressive faces and neutral faces in 2D. With the landmark coordinate differences between expressive faces and the neutral face of same person, the gender recognition rates for each expression with linear-kernel SVM classifier are: Happy (60.47\%), Disgust (59.68\%), Surprise (60.82\%) and Sad (64.82\%). With the differences of landmark distances between expressive and neutral faces, the gender recognition rates are: Happy (62.02\%), Disgust (55.38\%), Surprise (66.94\%) and Sad (62.96\%). The performances are even lower with the Random Forest classifier. In comparison with 3D counterpart, the expression based gender recognition performances on the 2D landmarks are much lower and ineffective in Happy and Disgust expressions. Recall that the experiments are performed on the same subjects and scans captured at the same instant, the only difference lies in the feature modality, 2D or 3D. These results demonstrate the advantage of using 3D mesh, in capturing gender related facial morphological changes during expression. In summary, it's demonstrated that the 2D landmarks based approaches are effective in capturing gender related cues in high-level facial structure, but not as effective in describing more fine-grained facial differences, such as the detailed morphological changes during facial expressions which also incorporate gender patterns. 

\begin{table}[]
\centering
\caption{Gender recognition using 2D Landmarks Coordinates and Landmark Distances using Linear-kernel Support Vector Machine. }
\vspace{0.5cm}
\label{tab:gender_all_2D}
\begin{tabular}{llllll}
\hline
\multicolumn{1}{|l|}{Feature} &\multicolumn{1}{|l|}{Classifier} & \multicolumn{1}{l|}{Female} & \multicolumn{1}{l|}{Male} & \multicolumn{1}{l|}{ALL}   \\ 
\hline
\multicolumn{1}{|l|}{Coordinates} & \multicolumn{1}{|l|}{SVM} & \multicolumn{1}{l|}{84.89\%} & \multicolumn{1}{l|}{87.58\%} & \multicolumn{1}{l|}{86.31\%}   \\ 
\hline
\multicolumn{1}{|l|}{Distances} & \multicolumn{1}{|l|}{SVM} & \multicolumn{1}{l|}{86.23\%} & \multicolumn{1}{l|}{87.88\%} & \multicolumn{1}{l|}{87.07\%}   \\ 
\hline

\end{tabular}
\end{table}


\subsection{Comparison with related studies}

In this study, we have achieved gender recognition rate of 79.15\% and 72.15\%, using only the 3D morphological changes presented in the Happy and Sad expressions. The methodology requires only a neutral scan and an expressive scan to characterize the 3D shape changes in expression. In comparison, the work presented in \cite{bilinski2016} studied video-level gender recognition from facial expression, where a combination of 5 different types of 2D shape and appearance features were extracted from each frame of the video to learn a fisher vector embedding representation for gender recognition. Although they claimed a relatively higher gender recognition rate of 86\% - 91\% on the evaluated 2D video dataset in 5 fold cross-validation, their model incorporates high computational complexity and lacks in interpretability. In comparison, the 3D based method in this work establishes a much more intuitive and robust way of capturing 3D facial deformations during expressions. In comparison with \cite{dantcheva2017}, this work improves the accuracy of expression based gender accuracy by about 13\%. We have also studied more expressions than \cite{bilinski2016,dantcheva2017}, and reported their effectiveness in revealing gender.

\subsection{Discussions}

In this section, we experimented with the 3D shape features for expression based gender recognition. Experimental results demonstrate that gender can be recognized with considerable accuracy through 3D shape changes in Happy and Disgust expressions, but not in Surprise and Sad expressions. Through PCA and t-test analysis, we demonstrated that male and female expressions are characterized differently, and gender related cues exhibit differently in terms of location and magnitude during facial expressions. In comparison with the 2D methods, the 3D based approach is demonstrated to be more effective in capturing gender related patterns from expressions.

\section{Conclusion}

In this work, we proposed to investigate the problem of gender recognition from facial expression, in particular with 3D face modality. Firstly, through expression-specific evaluation, we demonstrate that 3D gender perception with machine vision technique receive considerable impact from facial expressions. The gender recognition accuracy improves when training and testing within scans from same expression, which effectively reveal the strong correlation between perceived gender and performed expression. In further, we extract features capturing 3D facial deformations during each facial expression, and perform gender recognition experiments with these facial motion cues. Experimental results show that gender could be recognized with considerable accuracy from the 3D shape deformation occurred during Happy and Disgust expressions. While Surprise and Sad expressions don't have much strength in gender classification when studying the motion feature alone. 

The work also presents some limitations. As we use the difference of two 3D scans (a neutral face and an expressive face) to extract morphological changes during facial expression, we the full shape dynamics during facial expression, which could result in the loss to the incorporated gender information during facial expression. Recall that for human observers, the accuracy for recognizing gender from motion is much higher \cite{zibrek2013}. There is high potential that the gender recognition accuracy improves when studying the whole expression dynamics. Also, we are working on posed expressions, which could possibly different from the real spontaneous expressions, although \cite{dantcheva2017} showed that posed (deliberate) and spontaneous smiles do not differ much in the facial dynamic features. Limited by the available dataset, only the first limitation could be tackled in the near future, for which we plan to work on 3D facial expression videos on the BU4D dynamic 3D facial expression dataset \cite{yin2008}, to reveal the effectiveness of using fine-grained facial dynamics for gender recognition in 3D. We are also interested in exploring facial expressions cues for recognizing other biometric traits, such as Ethnicity, and Age \cite{dibekliouglu2015}.

\begin{IEEEbiographynophoto}{Baiqiang XIA} is currently a Postdoctoral Research Associate at the School of Computing and Communications, Lancaster University, United Kingdom. He was a research fellow in school of computer science and statistics in Trinity College Dublin, Ireland during 2015-2016. He received his PhD in 2014 in University of Lille 1 - Science and Technologies, Lille, France, in the area of 3D facial soft-biometrics recognition. He was the recipient of the Best Paper Award in the area of Image and Video Understanding in VISAPP 2014. His research interests are in face recognition, facial soft-biometrics recognition, hand motion recognition, and machine learning.

\end{IEEEbiographynophoto}

\end{document}